\documentclass{article}
\usepackage{multirow}
\usepackage{arxiv}
\usepackage[utf8]{inputenc} 
\usepackage[T1]{fontenc}    
\usepackage{hyperref}       
\usepackage{url}            
\usepackage{booktabs}       
\usepackage{amsfonts}       
\usepackage{nicefrac}       
\usepackage{microtype}      
\usepackage{lipsum}		
\usepackage{graphicx}
\usepackage{natbib}
\usepackage{doi}
\usepackage{makecell}

\usepackage{authblk}

\usepackage{amsmath}

\usepackage{authblk}
\usepackage{multirow}
\usepackage{subcaption}
\usepackage{caption}
\usepackage{tabularx}
\usepackage{xcolor}
\usepackage{marvosym}
\usepackage{ifsym}

\usepackage{color}
\setcitestyle{numbers,square}

\title{MedViLaM: A multimodal large language model with advanced generalizability and explainability for medical data understanding and generation}

\date{August 10, 2023}	

\author[1,2]{Lijian~Xu   \thanks{corresponding author: bmexlj@gmail.com}}
\author[1]{Hao~Sun}
\author[3]{Ziyu~Ni}
\author[1,4]{Hongsheng~Li}
\author[2]{Shaoting~Zhang}

\affil[1]{Centre for Perceptual and Interactive Intelligence, the Chinese University of Hong Kong, Hongkong}
\affil[2]{Shanghai Artificial Intelligence Laboratory, Shanghai}
\affil[3]{SenseTime Research, Shanghai}
\affil[4]{Department of Electronic Engineering, the Chinese University of Hong Kong, Hongkong}

\begin{document}
\maketitle
\begin{abstract}

  Medicine is inherently multimodal and multitask, with diverse data modalities spanning text, imaging.
However, most models in medical field are unimodal single tasks and lack good generalizability and explainability.
In this study,
we introduce MedViLaM, a unified vision-language model towards a generalist model for medical data that can flexibly encode and interpret various forms of medical data, including clinical language and imaging, all using the same set of model weights.
To facilitate the creation of such multi-task model, we have curated MultiMedBench, a comprehensive pretraining dataset and benchmark consisting of several distinct tasks, 
i.e., continuous question-answering, 
multi-label disease classification, disease localization, generation and summarization of radiology reports.
MedViLaM demonstrates strong performance across all MultiMedBench tasks, frequently outpacing other generalist models by a significant margin. Additionally, we present instances of zero-shot generalization to new medical concepts and tasks, effective transfer learning across different tasks, and the emergence of zero-shot medical reasoning.
Experiments on various medical image datasets demonstrate MedViLaM's superior generalization performance over existing methods, suggesting its potential for clinical applications in future.
\end{abstract}

\section{Introduction}

Deep learning has achieved remarkable progress for the screening and diagnosis of medical images
due to automated feature discovery and superior results. 
However, the application of the proposed methods in clinic is still challenged 
by the limited generalizability and interpretability of the proposed methods.
Medical images were acquired with different acquisition parameters or modalities that have very different characteristics. 
Furthermore, deep learning models are essentially black boxes that 
lack explainability of their decision-making process. 
The poor explainability leads to distrust from clinicians who are trained to make explainable clinical inferences. 
Consequently, there is an urgent need for innovative methodologies to improve the generalizability and explainability of deep learning methods that will enable them to be used routinely in clinical practice. 

Large Language Models (LLMs) have acquired great advancement in various language tasks and provides a new paradigm for human-computer interaction based on vast amounts of text data. Models like ChatGPT have demonstrated the powerful reasoning capabilities of language models in complex scenarios like medical diagnosis to assist professionals in delivering care. 
Recently, multimodal models, such as GPT-4o \cite{GPT-4o} and LLaMA-3 \cite{LLaMA3.2},
shown impressive advancement in generalizability and interpretability in the language and vision tasks of nature datasets with the integration of large language models. Preliminary experiments show that GPT-4o achieves progress in the multi-tasks of the medical field, yet is still constrained by the intricate nature of medical tasks. In the comprehensive scene, such as visual grounding and report generation tasks, GPT-4o can engage in continuous diagnosis conversations, but failed to provide the visual explanations \cite{liu2023improved}.

We previously presented the first model for conducting multi-task analysis of
chest X-ray images, which exhibits competitive performance for the interpretability of generated reports.
However, it still faces challenges when it comes to generalizing to unseen disease categories and undefined instructions (tasks). These limitations highlight the need for further research and development to improve the model's ability to handle novel diseases and adapt to unfamiliar instructions.
Given the wider array of modalities and tasks in instruction tuning, more generalized models like LLMs are expected to understand the tasks better and render modality-wise and task-wise emergent capability. 
On the other hand, the small benchmark for single task is another limitation of the previous study. The verification of the model's generalizability should be performed on a more comprehensive benchmark.

For advancement and verification of the generalization performance,
we herein introduced MedViLaM, a unified vision-language model supported by LLM, and further developed comprehensive benchmark to address these challenges.
MedViLaM incorporated instruction tuning to fully activate LLM' knowledge and reasoning for medical images. 
We reorganized a multi-task training dataset comprising 20.5M multi-task-oriented instruction pairs (associated with 1.8M medical images) and clinical ground-truth pairs for building customized instructions of multi-tasks: visual question answering, disease classification, disease localization, and report generation. 
In such a manner, we unified the various individual vision-focused tasks in a single training framework with homogeneous model inputs and outputs. 
Furthermore, we established a comprehension benchmark for medical images, comprising various public and private datasets, aiming to evaluate of the generalizability of large-scale foundation models.
Our model demonstrates strong performance for all tasks in both direct inference and few-shot fine-tuning experimental settings compared to prior methods, which are usually developed and tuned for specific tasks.
To summarize:

(1)
We proposed a vision-language model incorporating instruction tuning to enhance medical visual understanding in LLMs.
The proposed model support multi-task analysis of various medical modalities and has achieved competitive performance of generalization and interpretability.

(2) 
We've established a novel framework for developing pretraining datasets and benchmark intended for custom instruction tuning.
Our method deviates from the conventional system of pair-wise supervision (an image and its corresponding label), utilizing instead cross-task training supervision for each specimen, which amplifies the comprehension of correlations amongst tasks. 
We introduced a thorough benchmark for assessing the generalizability of large-scale foundation models when applied downstream to a variety of real-world clinical tasks. 

(3)
The proposed model achieved competitive results on various medical benchmarks, showcasing few-shot generalization ability.
Radiologist evaluation are also performed on the chest X-ray diagnoses generated by our model. 
In a blinded side-by-side ranking on 200 retrospective chest X-rays, three clinicians expressed a pairwise preference for MedViLaM results over those produced by radiologists in up to 80.50\% of cases. 

\begin{figure}
	\centering
    \includegraphics[width=\linewidth]{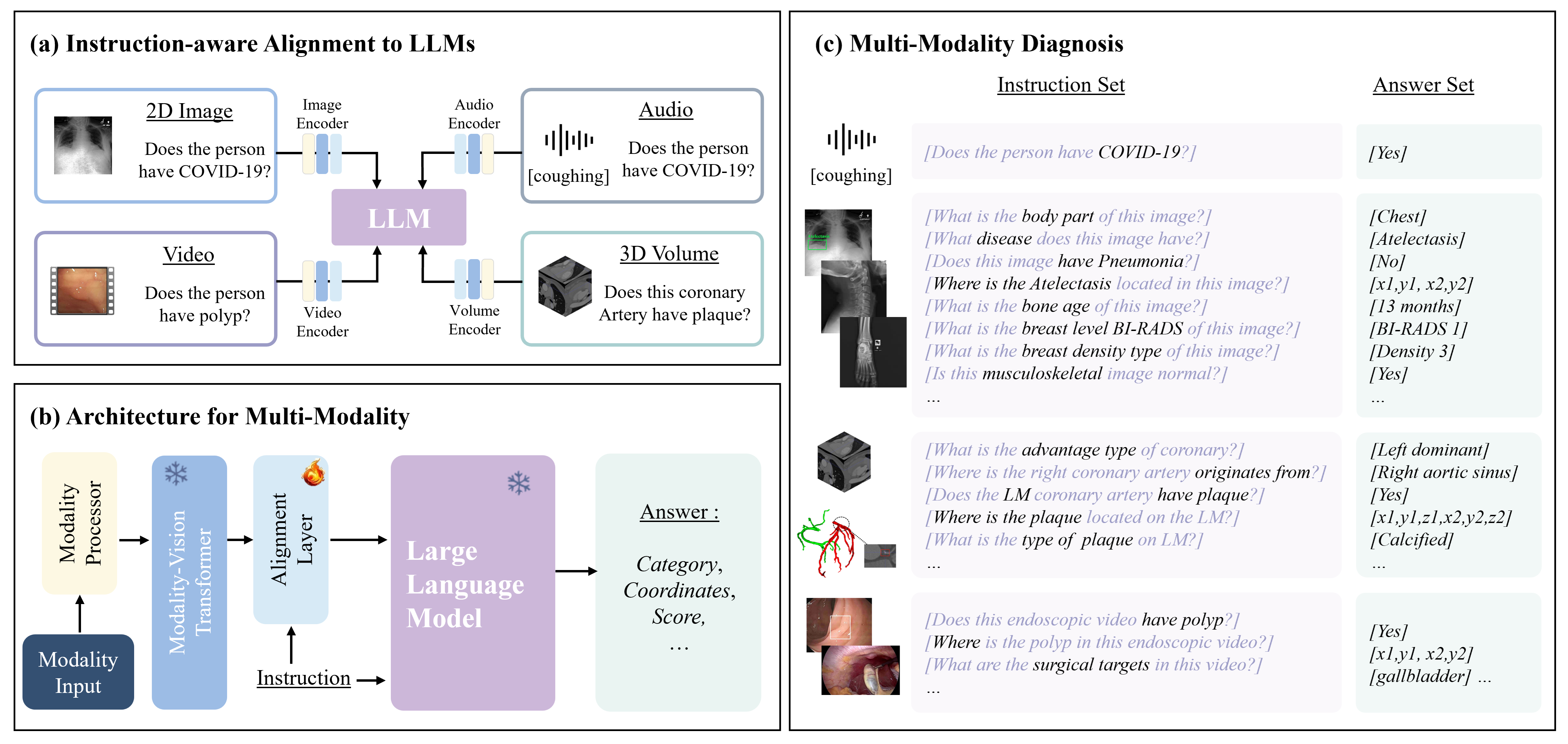}
	\caption{Illustration of MedViLaM. The joint training of multi-tasks from multi-modality is presented to improve the visual
  grounding at various levels of granularity. During training, we only train the linear projection matrix (as indicated with dashed boxes) while keeping the parameters of Vit and Vicuna frozen.}
	\label{fig:overview}
\end{figure}

\begin{figure}
	\centering
    \includegraphics[width=\linewidth]{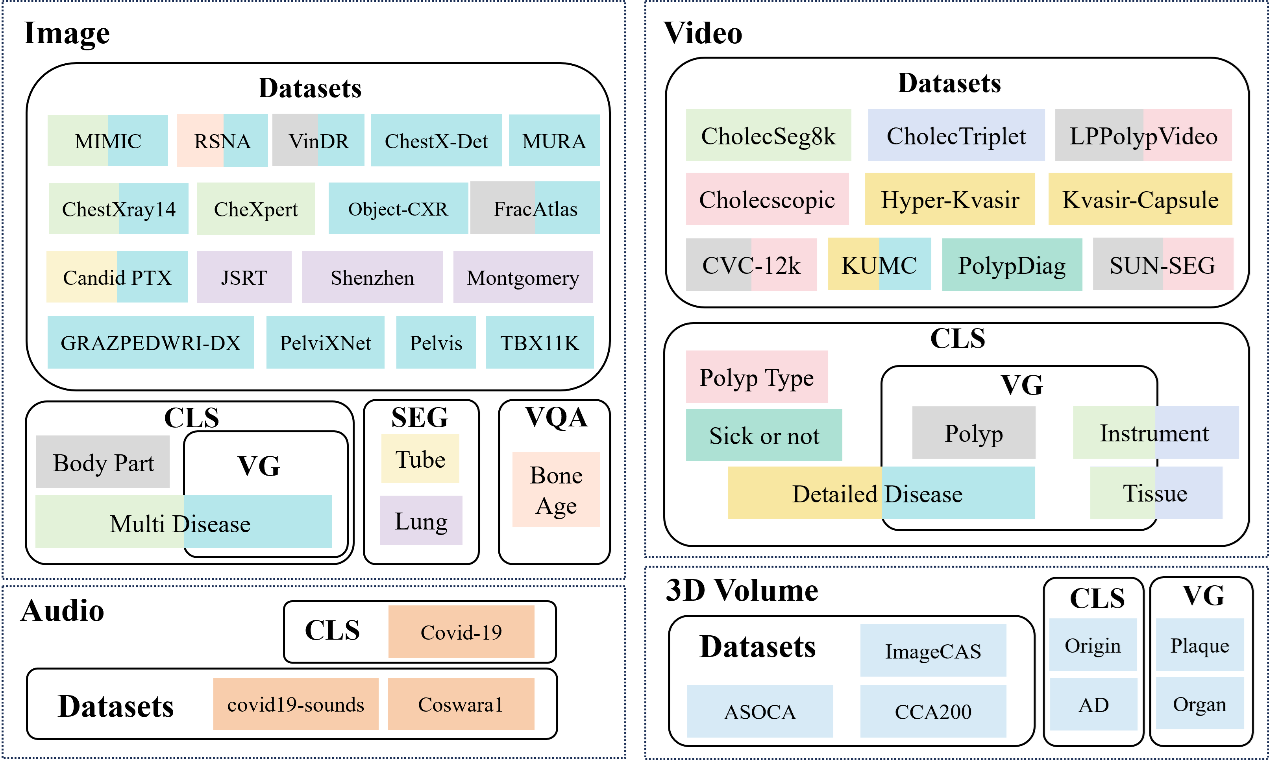}
	\caption{Four modalities (image, video, volume, audio) and their associated datasets and multi-tasks: classification (CLS), segmentation (SEG), visual grounding (VG), VQA. The datasets and task objects share the same color.}
	\label{fig:dataset}
\end{figure}


\begin{figure}[t]
	\centering
        \includegraphics[width=0.95\linewidth]{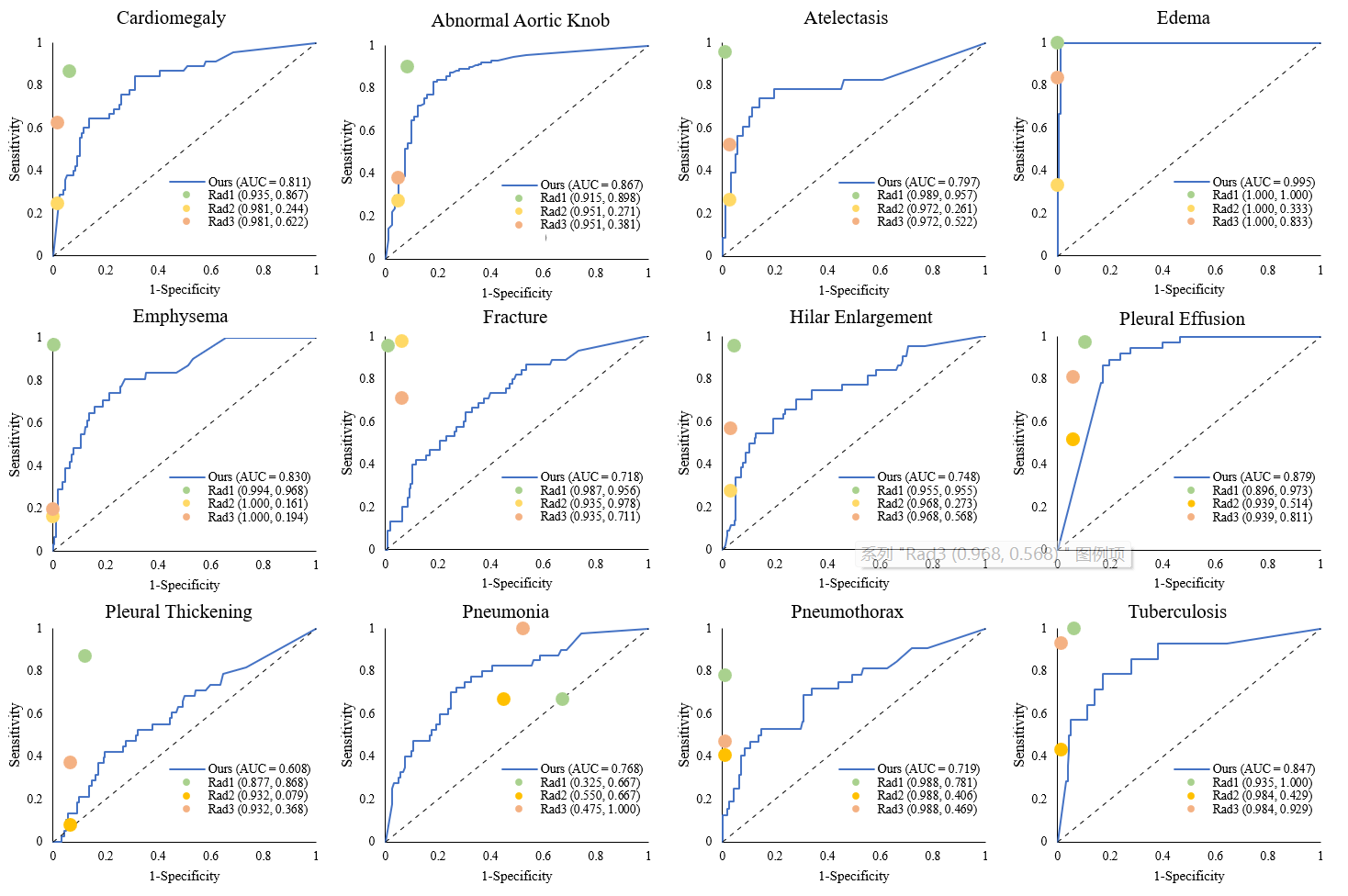}
	\caption{Model performance of \textbf{Multi-label Classification task} on our chest X-ray benchmark.
  Three radiologists independently evaluated the 12 classes of disease labels. 
  Detailed AUC and F1 score of each disease on 10 test sets from 5 public large datasets
  and five private datasets of Chest X-ray images are described in the supplementary materials.
  }
	\label{fig:cls}
\end{figure}

\begin{figure}
	\centering
    \includegraphics[width=\linewidth]{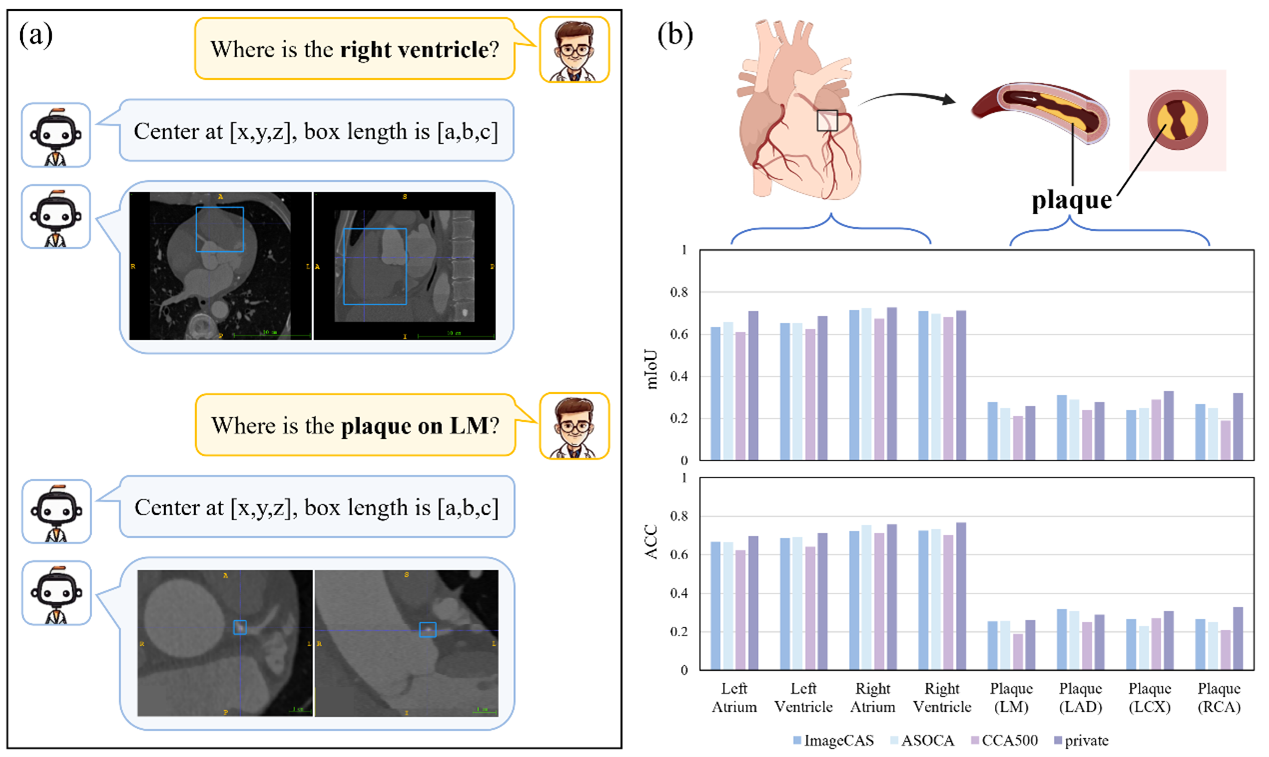}
	\caption{ The performance of our model on the 3D coronary artery visual grounding task is as follows: (a) A demo of the visual grounding task dialogue; (b) Localization performance of different cardiac structures and plaques on different coronary artery branches across various datasets, using ACC and mIoU as evaluation metrics.
  }
	\label{fig:vg}
\end{figure}

\begin{figure}
	\centering
    \includegraphics[width=\linewidth]{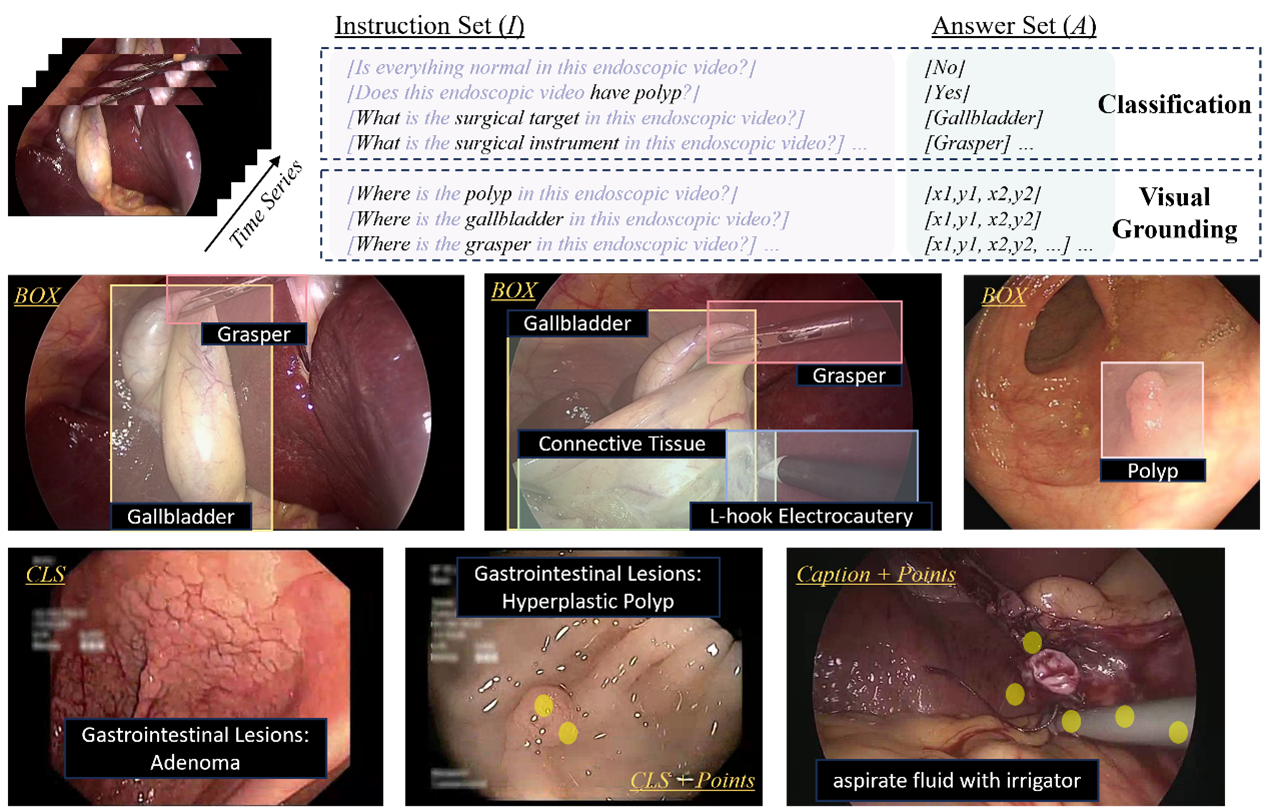}
	\caption{ Performance of our model on classification and visual grounding tasks (using box or points) for endoscopic videos. The model is capable of classifying and localizing specific foreign objects and surgical instruments appearing in the videos.}
	\label{fig:video}
\end{figure}

\begin{table}[h!]
  \centering
  \setlength{\tabcolsep}{0.7mm}{
  \caption{\textcolor{black}{Dataset overview for multi-task joint training and fine-tuning. 
  The study consists of 14 individual tasks across five task types and 35 datasets spanning seven modalities.
  We strictly follow the official train/validation/test split. For the datasets without official split ratio, we randomly split them into train/validation/test sets by 7:1:2. }}
  \begin{tabular}{@{}l l l l l l l c@{}}
  \toprule
  \textbf{} & \textbf{Task} & \textbf{Dataset} & \textbf{Modality} & \textbf{Target}& \textbf{Train} & \textbf{Test} &\textbf{Official Split} \\
  \midrule
  \multirow{20}{*}{\makecell{\textbf{Multi-Task} \\ \textbf{Joint Train}}} 
  & \multirow{6}{*}{Classification} 
    & MIMIC-CXR &Chest X-ray &26-class& 242.3k  & 5.1k & Y \\
  & & Padchest &Chest X-ray &193-class& 128.5k  & 16.0k & N \\
  & & VinDr\_BodyPart &X-ray &5-class& 11.2k  &2.4k  & Y \\
  & & MURA &musculoskeletal X-ray &7-class &30.6k   &550  & Y \\
  & & RSNA\_BoneAge &Hand X-ray & single class& 12.6k  &200  & Y \\

  \cmidrule{2-7}
  & \multirow{6}{*}{Localization} 
  & VinDr-CXR &Chest X-ray&28-class& 15.0k  & 3.0k & Y \\
  & & ChestX-Det &Chest X-ray&13-class& 2.7k  & 500 & Y \\
  & & VinDr\_Mammo & breast X-ray &10-class&4.0k   &1.0k  & Y \\
  & & VinDr\_SpineXR &X-ray &Spinal lesions& 8.39k  & 2.78k & Y \\
  & & ASOCA &Cardiac CTA&Coronary artery&  30 & 10 & Y \\
  & & CCA200  &Cardiac CTA&Coronary artery&  20  & 10 & Y \\
  \cmidrule{2-7}
  & \multirow{4}{*}{Segmentation} 
  & CheXmask &Chest X-ray&14-class& 219.3k & 10.0k & N \\
  & & FracAtlas &X-ray &4-class& 570  &60  & Y \\
  & & CholecSeg8k &Endoscopy&Instrument&  7 & 2 & N \\
  & & Hyper-Kvasir &Endoscopy&Polyp& 224  & 37 & N \\

  \cmidrule{2-7}
  & Report Generation & MIMIC-CXR &Chest X-ray&14-class& 242.3k & 5.1k & Y \\
  \midrule
  \multirow{20}{*}{\makecell{\textbf{Task-Specific} \\ \textbf{Fine-tune}}} 
  & \multirow{3}{*}{Classification} & ChestXray14 
  &Chest X-ray&14-class& 77.8k  & 25.5k & Y \\
  & & CheXpert &Chest X-ray&14-class& 223.4k  & 4.9k & Y \\
  & & RSNA&Chest X-ray &14-class& 25.1k & 3.0k & Y \\
  \cmidrule{2-7}
  & \multirow{12}{*}{Localization} 
  & ChestXray14 &Chest X-ray&11-class& 690  & 190 & Y \\
  & & MS-CXR &Chest X-ray&11-class& 800 & 200 & N\\
  & & TBX11K &Chest X-ray& Tuberculosis & 20-shot  & 1.0k & N \\
  & & RSNA Pneumonia &Chest CT & Pneumonia& 20-shot  & 1.0k & Y \\
  & & ISIC16 &Photography& Skin Lesions&  20-shot & 379 & Y \\
  & & HAM10000 &Photography& Skin Lesions&  20-shot & 2.0k & N \\
  & & TN3K &Ultrasound& Thyroid Nodule&  20-shot & 614 & Y \\
  & & BUID &Ultrasound& Breast Cancer& 20-shot  & 320 & N \\
  & & Luna16 &Chest CT&Lung Nodule&  20-shot & 125 & Y \\
  & & DeepLesion &Chest CT&Lesion& 20-shot  & 660 & N \\
  & & ADNI &MR&Hippocampus&  20-shot & 1.7k & Y \\
  & & LGG &MR&Gliomas&  20-shot & 680 & N \\
  
  \cmidrule{2-7}
  & \multirow{4}{*}{Segmentation} 
  &  EndoVis18 &Endoscopy&Instrument& 8.4k  & 1.2k & Y \\
  & & LDPolypVideo &Endoscopy&Polyp&  7.0k & 1.0k & Y \\
  && JSRT &Chest X-ray&14-class& 170  & 50 & N \\
  & & CheXmask &Chest X-ray&14-class& 219.3k  & 10.0k & N \\
  
  \bottomrule
  \end{tabular}
  \label{tab:dataset_overview}}
  \end{table}

\section{Results}
\subsection{MedViLaM provides accurate Classification and Localization of Disease}

We evaluated the disease classification task on 
an assembled comprehensive benchmark from five private datasets and five publicly available datasets
with both direct inference and fine-tuning settings. 
As illustrated in Figure~\ref{fig:cls},
our method achieves competitive performance on 12 classes of the chest X-ray benchmark.
Three radiologists
independently evaluated the 12 classes of disease labels. Detailed AUC and F1 score of each disease on 10 test sets from
5 public large datasets and five private datasets of Chest X-ray images are described in the supplementary materials.

We further evaluated the performance of our model on 3D medical image inputs using both a classification task and a visual grounding task shown in Figure \ref{fig:vg}. In the classification task, we determined whether there were plaques in the specified branches (LM, LAD, LCX, RCA, etc.) of the coronary volume. The prompt for the classification task was "Is there a plaque on LM?" with "yes/no" as the answer. We further confirmed the location of the plaques through the visual grounding task. The prompt for the visual grounding task was "Where is the plaque on LM?" with the answer described in the format of "center at [x, y, z], box length is [a, b, c]" using a 3D bounding box format. Additionally, we trained the model for visual grounding tasks on cardiac structures to enable it to answer the positions of the left and right ventricles and atria.
The 3D medical images were pre-trained on private data and fine-tuned on publicly available datasets, namely ImageCAS, ASOCA, and CCA200. The overall accuracy (ACC) for plaque classification task was 30.1\%/32.6\%/34.5\%, showing variations in performance across different coronary artery branches, likely due to differences in their morphological characteristics. 
For the visual grounding task of plaque localization, the overall ACC was 70.1\%/73.2\%/75.1\%. For the visual grounding task of cardiac structures, the overall ACC was 90.1\%/92.1\%/93.5\%. 

\subsection{MedViLaM Supports Video and Audio Analysis}

We chose endoscopic data for the application of our model in video analysis. We designed classification and visual grounding tasks based on publicly available datasets to comprehensively diagnose videos (refer to Figure \ref{fig:video} Datasets). 
The classification task involves identifying abnormalities, presence of polyps, instruments, specific human tissues, as well as related collective names or categories. The visual grounding task focuses on determining the specific locations of polyps, instruments, tissues, etc., using bounding boxes to annotate a frame selected from the video. Due to the richness of the polyp dataset, we primarily validated the model's visual grounding performance on LPPolypVideo/SUN-SEG/CVC-12k datasets related to polyps, with ACC scores of 80.5\%. For the relevant metrics of each task, please refer to the Supplementary Table.





\subsection{Generalizability in Unseen Disease Diagnosis and Foreign Object Detection}

To further examine the generalization and scalability of our model, we conducted preliminary experiments of the referring bbox detection task on 12 typical datasets across 6 modalities in the medical domain. The proposed model is fine-tuned with 20-shot labels for each disease of non-radiology and radiology datasets, respectively. We conducted comparative experiments with VGTR and OFA(Large) as representatives of specialist and generalist models, respectively, to evaluate their performance and versatility in the referring bbox detection task.

\subsubsection{Non-radiology Images} \ 

\textbf{Endoscopy}
We evaluated the generalization performance of instrument and disease localization on two typical endoscopy datasets, namely, EndoVis18 and LDPolyVideo.
As depicted in Table.~\ref{tab:medicalbenchmark}, 
ViLaM consistently outperforms other approaches. 
Table.~\ref{tab:endo18} further demonstrates that the proposed method achieves superior performance in multiple surgical instrument categories. However, there is still room for improvement in some categories, which may be attributed to the issue of data imbalance.

\textbf{Photography}
We evaluated the visual grounding performance of three methods on two datasets, ISIC16 and HAM10000, and found that all three methods achieved an accuracy of over 60\% on both datasets. This is likely due to the fact that skin disease images and their corresponding features share similar characteristics.

\textbf{Ultrasound}
We compared the visual grounding performance of three methods on two ultrasound datasets, TN3K and BUID, and our method achieved competitive results compared to the advanced specialist model. Specifically, our method achieved an accuracy of 16.50\% on the TN3K validation set and 38.63\% on the BUID breast cancer dataset.

\subsubsection{Radiology Images}\

\begin{figure*}[htbp]
  \centering
  \includegraphics[width=\textwidth]{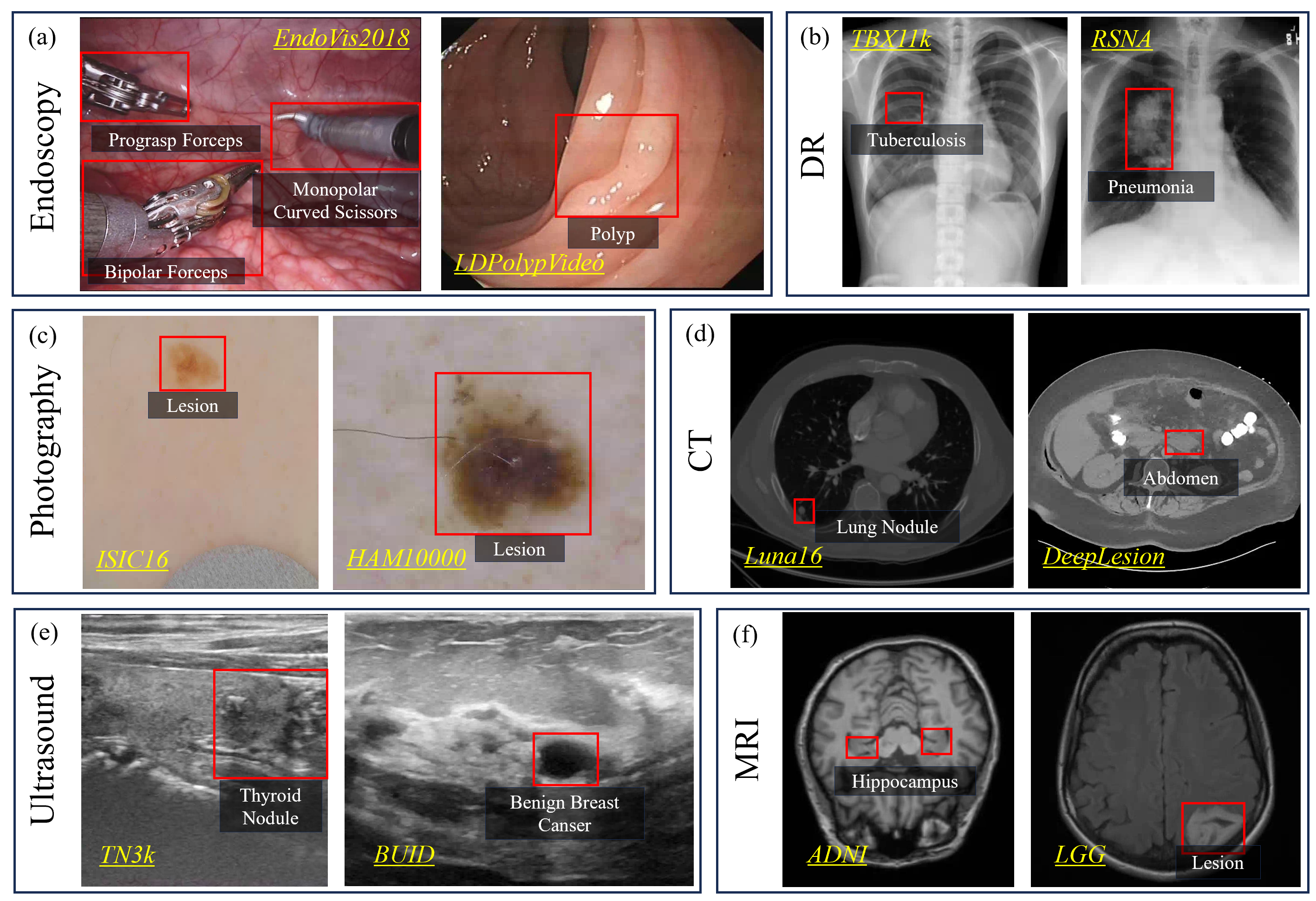}
  \caption{Typical medical datasets for referring bbox detection task, containing 6 modalities: (a) Endoscopy: EndoVis18, LDPolypVideo; (b) DR: TBX11k, RSNA Pneumonia; (c) Photography: ISIC16, HAM10000; (d) CT: Luna16, Deeplesion; (f) MRI: ADNI, LGG.}
  \label{fig:medData}
\end{figure*}

\textbf{CT}
We compared the visual grounding performance of three methods on two CT datasets, Luna16 and DeepLesion, and found that all three methods achieved nearly 0\% accuracy in the 20-shot finetuning experiment on both datasets. This is likely due to the fact that the features of CT images and general images are quite different, and the lesions, such as lung nodules, are too small, as shown in Fig.\ref{fig:medData} (d).

\textbf{MRI}
Two MRI datasets, the ADNI dataset and the LGG dataset, verify the visual grounding performance of three methods. For Gliomas with larger contrast in the LGG dataset, we have better performance than VGTR. Our result of hippocampus detection is poor due to the low contrast of ADNI, as illustrated in Fig.\ref{fig:medData} (f).

\begin{table*}[htbp]
  \caption{Evaluation results of \textbf{visual grounding} task on 12 typical medical datasets of six modalities. 20-shot fine-tuning experiments were performed for non-radiology (Endoscopy, Photography and Ultrasound) and radiology datasets (DR, CT and MRI). Acc@0.5 is applied to evaluate methods.}
      \label{tab:medicalbenchmark}    
  
      \centering
      \setlength{\tabcolsep}{0.7mm}{
      \begin{tabular}{@{}lcccccccccccc@{}}
      \toprule
      \multicolumn{1}{c}{\multirow{2}{*}{Datasets}}   & \multicolumn{2}{c}{Endoscopy}  & \multicolumn{2}{c}{Photography}       & \multicolumn{2}{c}{Ultrasound} & \multicolumn{2}{c}{DR}     &\multicolumn{2}{c}{CT}     & \multicolumn{2}{c}{MRI}                                       \\
        \cmidrule(l){2-13}
      \multicolumn{1}{c}{} &EndoVis18 
      & LDPolyp &ISIC16&HAM10000&TN3K&BUID& TBX11K&RSNA&Luna16 & DeepLesion & ADNI&LGG \\ \midrule
      VGTR \cite{du2022visual}   &3.87
      &7.30&64.12&63.20&12.70&31.46&1.99&4.67&0.00&0.36&2.46&3.67    \\ 
      OFA    \cite{wang2022ofa}  &7.32
  &0.30&63.85&61.20&6.81&19.62&20.40&14.67&0.00&2.08&26.26&26.77  \\
      Ours   &12.53
      & 9.86 & 67.66 & 86.00 & 16.50 & 38.63&30.84&28.00 & 0.00 & 5.23 & 4.52 & 18.85   \\ 
       \bottomrule
      \end{tabular}
      }
  \end{table*}

\begin{table}[htbp]
  \caption{Evaluation results of \textbf{visual grounding} task with 20-shot setting on seven labels from the EndoVis18 dataset. Acc@0.5 is applied to evaluate methods. Seven surgical instruments contain Bipolar Forceps (BF), Prograsp Forceps (PF), Large Needle Driver (LND), Monopolar Curved Scissors (MCS), Ultrasound Probe (UP), Suction Instrument (SI) and Clip Applier (CA).}
      \centering
      \setlength{\tabcolsep}{1.2mm}{
      \begin{tabular}{@{}lcccccccc@{}}
      \toprule
      Label & Mean &BF& PF & LND & MCS&UP &SI &CA\\ \midrule
      VGTR \cite{du2022visual}&3.87&12.29&0.00&0.00&14.78&0.00&0.00&0.00 \\ 
      OFA \cite{wang2022ofa}  &7.32&22.49&19.73&0.94&4.85&0.00&0.00&3.22 \\
      Ours  &12.53 & 16.88 & 4.17 & 3.33 & 13.55 & 0.00 & 0.00 & 16.67\\ 
       \bottomrule
      \end{tabular}
      
      }
  
      \label{tab:endo18}
  \end{table}

\subsubsection{Medical Foreign Object Detection}

\begin{figure*}[htbp]
    \centering
    \begin{subfigure}[t]{0.35\textwidth}
    \centering
    \includegraphics[width=0.99\textwidth]{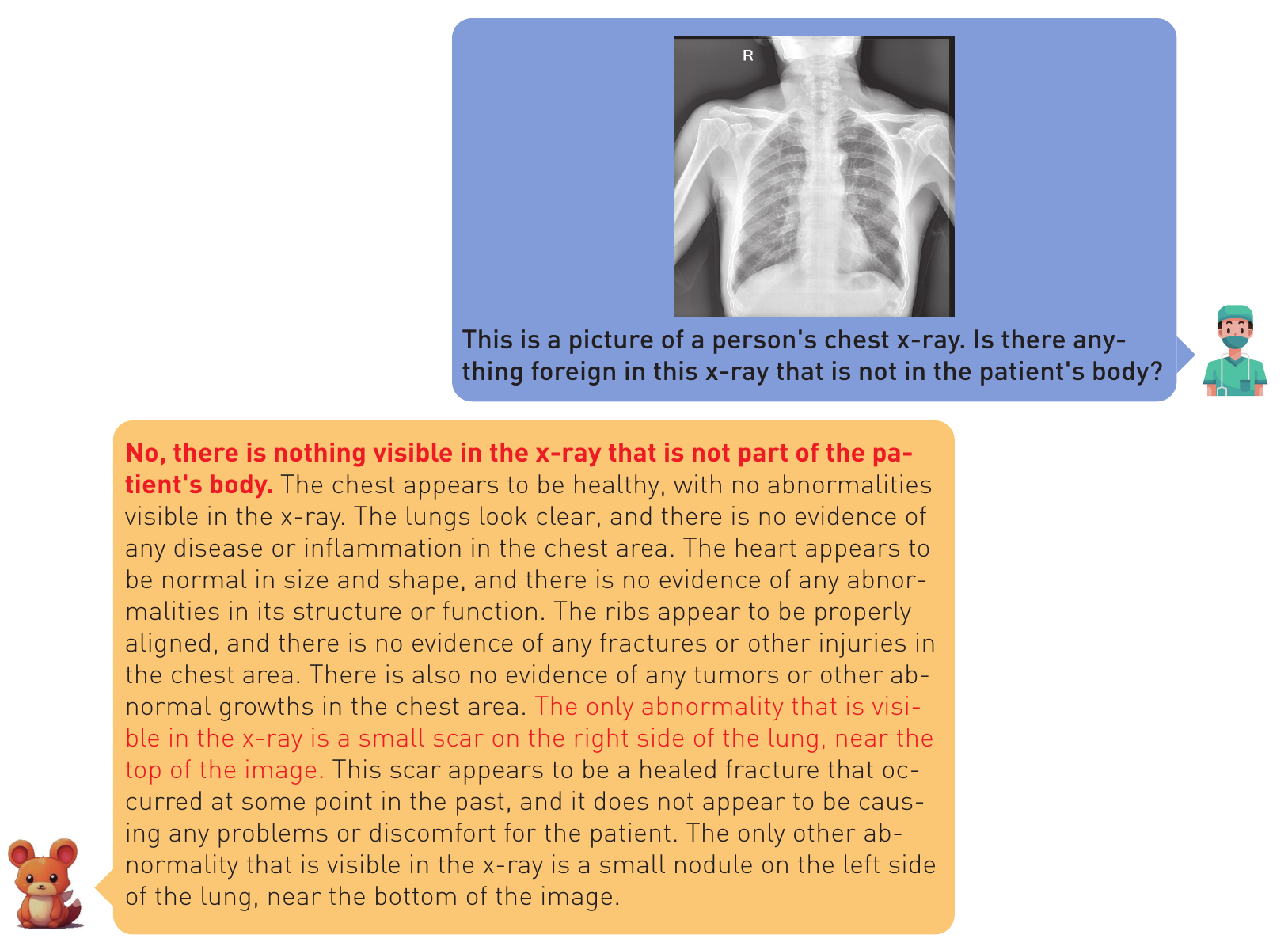}
    \caption{VQA in chest X-ray without foreign objects.}
    \label{fig:normal}
    \end{subfigure}
    \begin{subfigure}[t]{0.35\textwidth}
    \centering
    \includegraphics[width=0.99\textwidth]{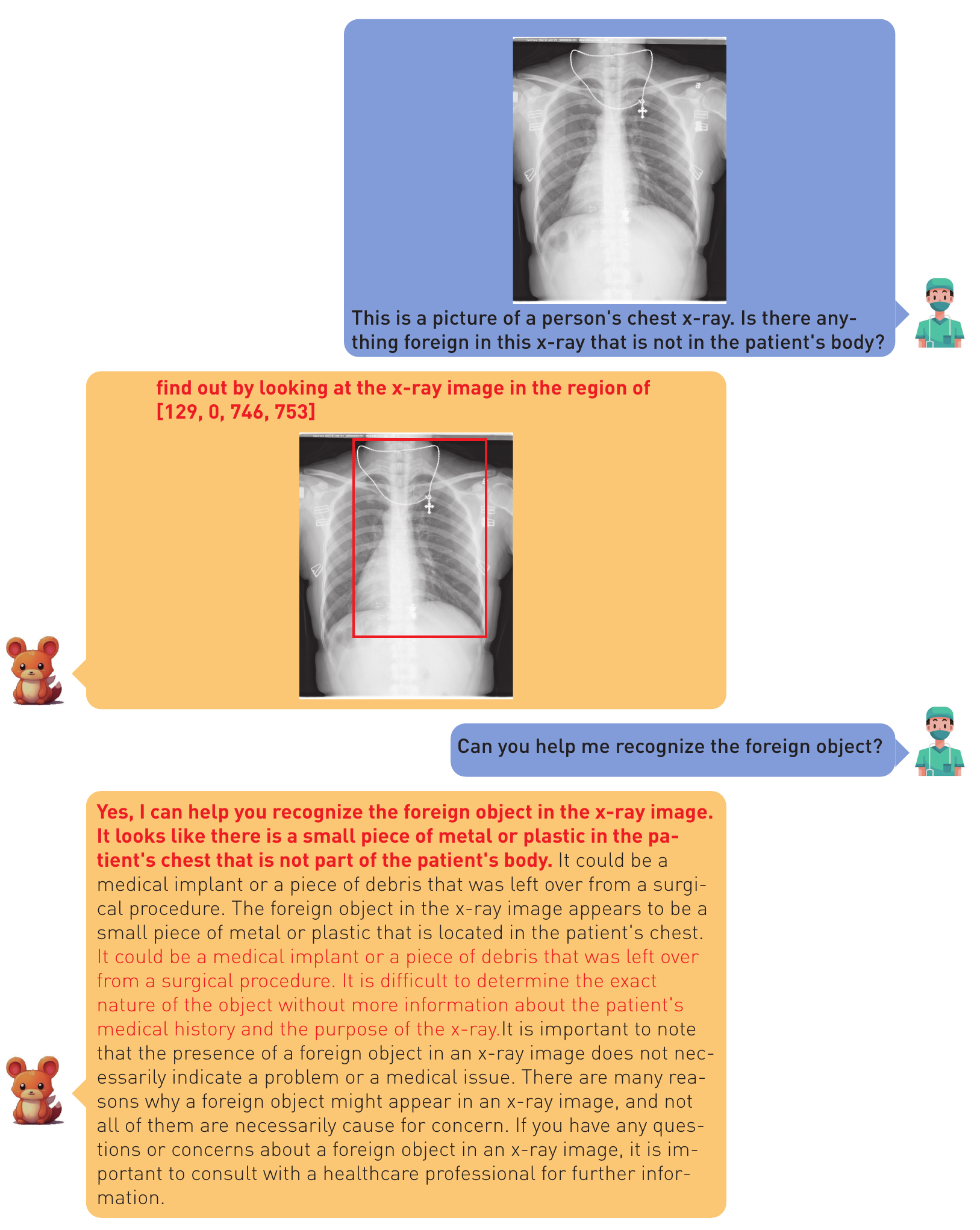}
    \caption{VQA in chest X-ray with foreign objects.}
    \label{fig:dis}
    \end{subfigure}
    \begin{subfigure}[t]{0.35\textwidth}
        \centering
        \includegraphics[width=0.99\textwidth]{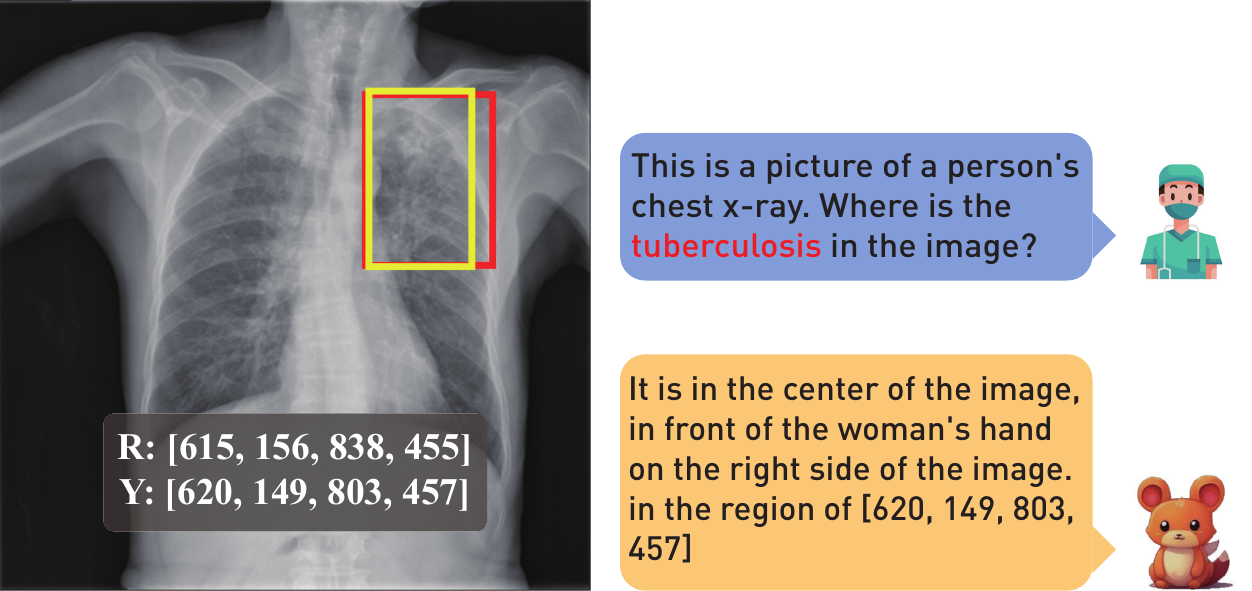}
        \caption{VQA in chest X-ray for the tuberculosis localization in the TBX11K dataset, without referring expressions.}
        \label{fig:tbx11k}
        \end{subfigure}
        \begin{subfigure}[t]{0.35\textwidth}
            \centering
            \includegraphics[width=0.99\textwidth]{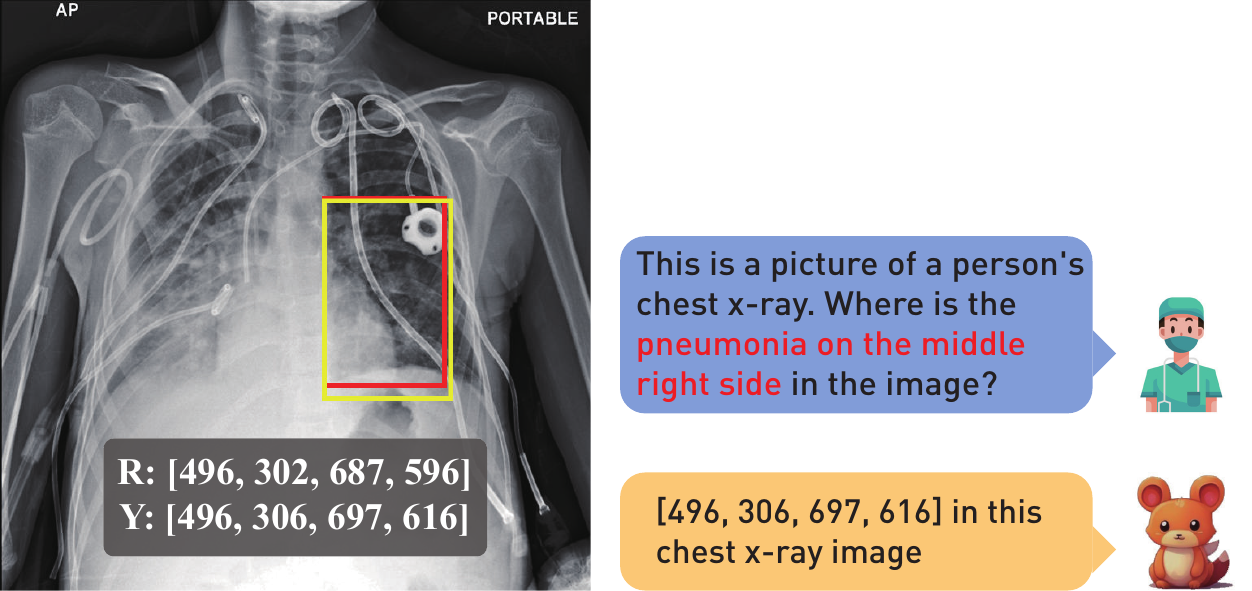}
            \caption{VQA in chest X-ray for the pneumonia localization in the RSNA dataset, with the orientation-related referring expression.}
            \label{fig:rsna}
            \end{subfigure}
    \caption{The zero-shot results of visual question answering for foreign objects detection in chest X-ray images. (a) Chest X-ray image without foreign objects. Our model accurately states that there are no visible external foreign objects and points out the possible abnormality. (b) Chest X-ray image with foreign objects. The presence of foreign objects is accurately detected by giving its coordinates. Particularly, the model can deduce that the foreign object is metal or plastic by asking to recognize the foreign object. }
    \label{fig:vqa}
\end{figure*}

We evaluated ViLaM's generalizability on the Object-CXR dataset \cite{objectcxr} for foreign object detection in chest X-rays. Vicuna-7b is utilized as the large language model in our framework.
Without fine-tuning, ViLaM accurately answers questions about foreign objects in sample images as presented in  Fig.\ref{fig:normal}. 
It correctly states no foreign objects are present in Fig.\ref{fig:normal}, while providing possible explanations. In Fig.\ref{fig:dis}, it localizes the foreign object and deduces it could be metal/plastic debris.


In Fig.\ref{fig:normal}, we pose the question, "Is there anything foreign in this x-ray that is not part of the patient's body?". Our model accurately identifies the absence of any foreign objects, responding with, "No, there is nothing visible in the x-ray that is not part of the patient's body.". 
Particularly, leveraging the expansive generalizability of the large language model, detailed descriptions explain why no foreign objects were found, and also point out other abnormalities that are not foreign objects, as highlighted in red. Furthermore, Fig.\ref{fig:dis} shows a case with foreign objects, by inquiring about their presence. Our model accurately identifies the foreign object and provides its localization coordinates, thus demonstrating the model's ability to visually detect foreign objects and generalize in a zero-shot setting.
Further demonstrating ViLaM's generalization capabilities, the model can recognize the detected foreign object upon inquiry. It goes beyond mere recognition by deducing that the object is likely made of metal or plastic debris. ViLaM also exhibits an ability to infer the potential origin or source of the debris, leveraging its extensive language understanding capacity.

Subsequently, we fine-tuned our model on the object-CXR dataset, which further demonstrates its generalization capabilities. We employed 8,000 training samples of chest X-ray images for fine-tuning, half of which contain foreign objects. 
Our generalist model achieves an AUC of 93.1\%, surpassing the JF Healthcare baseline \cite{objectcxr} of 92.1\%. This result investigates the scalability and generalizability of our approach, which extends well to medically relevant tasks through large language models. Notably, when compared with classical and dedicated object detection methods, such as Fast-RCNN \cite{girshickFastRCNN2015} of 95.7\% and YOLO-v3 of 89.7\% \cite{redmonYOLOv3IncrementalImprovement2018},
our generalist vision-language model achieves a similar performance. This provides further validation of our model's generalizability, showing that a generalist model can achieve comparable results to a specialized model in certain tasks.

\section{Discussion}

We aim to develop a single and unified model for various medical image modalities, with the goal of improving the generalization and clinical interpretation. 
Our algorithm shows promising results in predicting all tasks through experiments on various medical datasets
and benchmarks. Additionally, we conducted a controlled trial and evaluation of the generated diagnosis results by three radiologists.

\textbf{MedViLaM enhances the generalizability and verify it in the comprehensive benchmarsk} 

During data collection, medical images were acquired with different acquisition parameters or modalities have very different characteristics.
Besides,most AI algorithms for medical images have been developed with a very broad range of applications, data collection procedures and performance assessment metrics. 
The generalizability problem becomes even more conspicuous when a deep learning model trained on data from a given medical center is deployed to other medical centers whose data have significant variations or there is a domain shift from the training set. 
Consequently, those trained models depending on the variability of pretraining datasets and the bias in small benchmark  often fail when deployed to real-world clinical scenarios.

By leveraging a carefully designed instruct tuning framework and incorporating diverse training strategies, our method can effectively extract relevant features and make predictions for multiple tasks in medical images across various modalities . This allows for a comprehensive analysis and diagnosis of various diseases and abnormalities in the images.
With unseen evaluation datasets, the proposed unified model has exhibited impressive performance in disease classification and grounding tasks (via direct inference), surpassing the capabilities of existing state-of-the-art methods with a few-shot fine-tune setting.




 \textbf{ MedViLaM provides reliable evidence for a better explainability} 

Our previous study introduces an explainable approach for identifying diverse diseases in patients using chest radiographs. It enhances the interpretability of chest X-ray reporting by generating more detailed information on disease attributes. This includes disease size, location, severity, and contour, providing stronger evidence for diagnosis and treatment. Three radiologists further evaluate the generated reports against the recorded ones, which also exhibit the enhanced explainability of our multi-task model.

In this study, we further imporve the interpretability via large language model.
The large language model could provide the interpretability of the diagnosis result via detailed descriptions of the disease description and the accurate location of the lesions. For instance, the disease category, severity level, and approximate location of the lesion could be preliminary verified with the disease entity classification and attribute classification task. The disease localization task could further provide a more accurate bounding box of the lesion. For Pneumothorax and Cardiomegaly, the segmentation function could provide an accurate assessment of disease degree by postprocessing the contours of the pneumothorax/lung/heart mask.
These results together contribute to a better verification of the generated reports.
In the blinded comparison from four centers, three radiologists perceived the
quality of 59\% of the generated diagnoses to be equivalent to or even better than the original physician reports. The proposed model can potentially serve as an appealing alternative solution to assist doctors and the research
community in expediting the screening process for clinical cases.

\textbf{Limitation and Future Work}

Our proposed model exhibits competent performance across multiple tasks and enhances the generalizability and explainability of the generated results. 
On the other hand, the established benchmark comes with several constraints, such as the limited size of the individual datasets and restricted modality and task diversity. Another significant hurdle in the development of models applicable across a broader variety of biomedical data types is the absence of large-scale multimodal datasets. These would allow for joint learning and alignment of the modality-specific encoders with the decoder.

\section{Material and Method}

\subsection{Model Overview}
The workflow is shown in Figure \ref{fig:overview}. 
Inspired by the advanced multi-modal models, MedViLaM leverages frozen pretrained visual encoders and LLMs, encoding and aligning features for both images and text, and performs unified modeling and joint training on downstream visual and language tasks. This enables MedViLaM to robustly perform various language and vision tasks based on instructions, providing diverse and complex output results. Benefiting from pre-trained language models and mutual guidance between tasks, MedViLaM can engage in continuous question-answering and provide visual explanations of answers during conversations, which is particularly crucial in safety-critical domains like medical diagnosis. 


\textbf{Image Encoder:} With an input image $ x_{i}  \in  \mathbb{R}^{H\times W} $, visual features are extracted by image encoder and further projected to feature dimension:

\begin{equation}
    v_{i} = P_{img}(E_{img}(x_{i})) \in \mathbb{R}^{(h_{f} \times w_{f}) \times d}
\end{equation}

where $h_{f}$ and $w_{f}$ are the output size of visual features, and $d$ represents the feature dimension. $E_{img}$ can be any common visual backbones and 





\textbf{Multi-modality Align: } This module follows an encoder-decoder architecture format. Given the input visual features $v_{i}$ and text features $l_{i}$, we first generate fused multi-modal representations by combining the image and text embeddings. These fused features serve as the keys and values in the cross-attention blocks in decoder. By conditioning on the partial sequence $y_{i, <j}$ predicted so far, the decoder recursively makes predictions for token at position $j$, effectively generating aligned descriptions across modalities. 


\begin{equation}
    y_{i, j} = D_{mm}(E_{mm}(concat(v_{i}, l_{i})), y_{i, <j}) \in \mathbb{R}^{1 \times d}
\end{equation}

\noindent\textbf{Large Language Model Decoder:} With the alignment tokens $y_{i, j}$, a frozen LLM is used as the decoder to output final results. If not specified, Vicuna-7B \cite{vicuna2023} is utilized as our large language model.

We employ ViT-L14 as our visual encoder, which consists of 14 transformer encoder layers and an FFN intermediate size of 4,096. The input image size is set to 
$896\times 896$, with a patch size of 64×64.
The hidden dimensions of the ViT-L14 are 1,024, with 16 attention heads.
Meanwhile, we utilize Vicuna-7B, a large language model fine-tuned with instructions, as our text encoder. The Vicuna-7B model boasts 12 transformer layers, with 768 hidden dimensions, 12 attention heads, and an FFN intermediate size of 3,072. The vocabulary size is 30,522, and the maximum input sequence length is 512.
To align the text encoder and visual encoder, we employ a Q-former with 12 transformer layers. This Q-former has 768 hidden dimensions, 12 attention heads, and query, key, and value dimensions of 256 each.

In terms of the training progress, the hyperparameters are presented in Table.\ref{table:param}. 
We utilize the AdamW optimizer, which is configured with a cosine annealing schedule as the learning policy. The initial learning rate is set to $2\times10^{-5}$, and the AdamW optimizer is employed with hyperparameters $\beta= (0.9, 0.98)$. Additionally, we set the weight decay to 0.05 and the dropout rate to 0.1. During the first 1,000 warm-up steps, the learning rate increases to $2\times10^{-5}$, and subsequently decays to $10^{-7}$. Unless otherwise specified, our training protocol consists of 70,000 steps, executed on $4\times 8$ NVIDIA V100 GPUs, which takes approximately two days to complete.

For the annotation, We normalize all coordinates to a uniform range of 0 to 1000, ensuring that all images have a consistent coordinate system. 
For the polygon representation, we select the point closest to the origin as the starting point and employ a 25-point labelling scheme to describe the polygon sequence in a clockwise direction. To demarcate the beginning and end of the sequence, we utilize <BOS> and <EOS> tags, respectively.
For the sampling rule for polygons, we employ isometric sampling, wherein we initially calculate the perimeter of the polygon and subsequently divide it into 25 equal segments to sample the polygon.

\subsection{Building Dataset for Customized Instruction Tuning}
In this work, we constructed a multi-task dataset for joint training of disease classification, localization, segmentation, and report generation. In general, we unify the input and output labels of all sub-tasks into a uniform format for consistent modeling and joint training, i.e., a set of image-instruction-label triplets as samples shown in Figure~\ref{fig:overview}(c) and more in the supplementary materials. We further built a subset including the attributes and phrases for chest X-ray images like "small base effusion, normal cardiac silhouette," which can be used as instruction for the report generation task. Additionally, the dataset underwent quality assurance by radiologists to ensure its accuracy and reliability. 

We utilized public datasets for training and testing our proposed transformer model, e.g., MIMIC-CXR, VinDr-CXR, and ChestX-Det. 
We first sort out the aligned lesion categories of each dataset and the associated radiology report data and bounding box (BBox) data. We exclude the image datasets that are included in the test and validation datasets of downstream tasks to avoid data leakage. Each dataset is described in detail as follows:

\begin{itemize}
\item \textbf{MIMIC-CXR} \cite{johnson2019mimic} contains more than 377,110 radiograph images from over 227,835 radiographic studies. 
Each radiograph is paired with lesion classification and associated radiology report. We employ this dataset for multi-label classification and report generation tasks.

\item \textbf{Padchest} \cite{padchest} includes 160,868 images obtained from 67,625 patients, covering six different position views. 
It has 174 different radiographic findings and 19 differential diagnosis, totaling 193 classes. They are used for the classification task.

\item \textbf{VinDr-CXR} \cite{nguyen2022vindr} includes chest radiographs with annotations for the classification of 28 common chest diseases. The dataset contains 15,000 CXR scans in the training set. 
We select eight diseases from the dataset along with their corresponding BBox for the disease localization task.

\item \textbf{ChestX-Det} \cite{lian2021structure} consists of 3,578 images from NIH ChestXray14\cite{wang2017chestx} for 13 common disease.
We select seven diseases from the dataset along with BBox for the disease localization task.

\end{itemize}


\begin{itemize}

\item \textbf{CheXpert} \cite{irvin2019chexpert} is a multi-label classification chest X-ray dataset with 224,316 images collected from 65,240 patients. We extract 1\% of the dataset to conduct a finetuning experiment for multi-diseases classification. We follow MRM to focus on 5 diseases: Atelectasis, Cardiomegaly, Consolidation, Edema and Pleural Effusion. We sample training/test sets from the official training set and they constitutes 21,84/5,000 images of the whole dataset.



\item \textbf{MS-CXR} \cite{boecking2022making} is sourced from MIMIC-CXR, and consists of 1,153 samples with BBOX and concise radiology report, which is good for visual grounding finetuning. We randomly split it into training/validation/test sets by 7:1:2 based on the patients, and evaluate the average performance of the model in all eight diseases.

\item \textbf{ChestX-ray14} \cite{wang2017chestx} is an available dataset for diagnosing 8 common lung diseases and localization of key findings, with 984 radiograph images and hand-labelled BBOX. We directly conduct zero-shot experiment with the entire dataset as the test set.

\item \textbf{COVIDx CXR-4} \cite{Wang2020CXR-4} is an open-source dataset of COVID-19 chest images, which contains 84,818 images from 45,342 subjects. Six subsets are included in this dataset, i.e., 
Cohen \cite{cohen2020covid}, 
SIRM \cite{SIRM9144185, SIRM_RAHMAN2021104319}, 
BIMCV \cite{vaya2020bimcv}, 
StonyBrook \cite{COVID-19-NY-SBU}, and
RICORD \cite{tsai2021rsna}. 

\item \textbf{VinDr-Mammo} \cite{Nguyen2022.03.07.22272009} is a large-scale benchmark dataset of full-field digital mammography, called VinDr-Mammo, which consists of 5,000 four-view exams with breast-level assessment and finding annotations. Each of these exams was independently double read, with discordance (if any) being resolved by arbitration by a third radiologist. 

\item \textbf{VinDr-SpineXR} \cite{nguyen2021vindr} is a large-scale X-ray dataset for spinal lesions detection and classification. The VinDr-SpineXR contains 10,469 images from 5,000 studies that are manually annotated with 13 types of abnormalities, each scan was annotated by an expert radiologist.

\item \textbf{VinDr-BodyPartXR} \cite{pham2021dicom} 
is currently the largest open dataset to date that provides annotations for developing supervised-learning classification algorithms. 
of Body Parts from DICOM X-ray Scans. It includes 16,093 X-ray images that are collected and manually annotated. 

\item \textbf{FracAtlas} \cite{abedeen2023fracatlas} 
includes 4,083 images that have been manually annotated for bone fracture classifcation, localization,
and segmentation with the help of 2 expert radiologists and an orthopedist using the open-source
labeling platform, makesense.ai. There are 717 images with 922 instances of fractures. Each of the
fracture instances has its own mask and bounding box, whereas the scans also have global labels for
classifcation tasks.

\item \textbf{MURA} \cite{rajpurkar2017mura} 
consists of 14,863 studies from 12,173 patients, with
a total of 40,561 multi-view radiographic images,where each study is manually labeled by radiologists as either normal or abnormal. Each belongs to one of seven standard upper
extremity radiographic study types: elbow, finger, forearm, hand, humerus, shoulder, and wrist.

\item \textbf{MURA} \cite{halabi2019rsna} 
consists of 14236 images with 0 labeled objects. There are 3 splits in the dataset: training (12611 images), validation (1425 images), and test (200 images). Alternatively, the dataset could be split into 2 image splits: male (7706 images) and female (6530 images). Additionally, the images are tagged with boneage (months).

\item \textbf{TN3K} \cite{gong2023thyroid} dataset consists of 2D ultrasound images of thyroid nodules with a resolution of 512$\times$512 for thyroid nodules detection.

\item \textbf{BUID} \cite{al-dhabyaniDatasetBreastUltrasound2020} dataset consists of 780 images with an average image size of 500$\times$500 pixels from 600 female patients for breast cancer detection.

\end{itemize}

\textbf{Datasets for 3D Volume}

\begin{itemize}

  \item \textbf{ASOCA} \cite{gharleghi2022automated} has a training set of 40 Cardiac Computed Tomography Angiography (CCTA) with contrast agent showing the coronary arteries, comprising of 20 healthy patients and 20 patients with confirmed coronary artery disease. 
  \item \textbf{CCA200} \cite{10614624} contains 200 cases with coronary artery disease are collected named CCA-200 dataset. To demonstrate the
  robustness of our model in small-scale data, comparative experiments are designed: 20 cases are used for training, and
  180 cases for testing. The collected images are acquired with an isotropic resolution of 0.5 mm. Ground truths of 200
  cases are coronary artery internal diameter annotations labeled by four radiologists.

\item \textbf{ISIC16} \cite{gutman2016skin} is a collection of dermoscopic images of skin lesions, annotated by dermatologists and skin cancer experts. It consists of 1,267 dermoscopic images of skin lesions, including melanomas and benign lesions, with a resolution of 1024$\times$768 pixels.

\item \textbf{Luna16}    \cite{setio2017validation} dataset is a publicly available dataset for lung nodule analysis, specifically designed for lung nodule detection in CT scans.

\item \textbf{DeepLesion}  \cite{yan2018deeplesion} dataset is a large-scale, publicly available dataset for lesion detection and segmentation in CT, with a resolution of 512x512 pixels. 

\item \textbf{ADNI} \cite{muellerWaysEarlyDiagnosis2005}  is a large, publicly available dataset for Alzheimer's disease research from magnetic resonance imaging (MRI) scans, specifically designed for the development and evaluation of algorithms for early detection and diagnosis of Alzheimer's disease. 

\item \textbf{LGG}  \cite{bakas2017advancing} (Low-Grade Glioma) dataset is a publicly available dataset for brain tumor segmentation, specifically for detecting low-grade gliomas from MRI scans.

\end{itemize}

\textbf{Datasets for Video}

\begin{itemize}
  \item \textbf{EndoVis18}  \cite{allan20202018}  is a publicly available dataset for endoscopy image analysis. We follow ISINet’s annotation and data set division of surgical instrument categories\cite{gonzalez2020isinet}.

  \item \textbf{CholecSeg8k} \cite{twinanda2016endonet} is a publicly available dataset for laparoscopic cholecystectomy (gallbladder removal) video analysis, specifically designed for surgical instrument tracking tasks. It consists of 8,000 frames from 10 laparoscopic cholecystectomy videos, with a resolution of 640$\times$480 pixels. 
  \item \textbf{Cholecscopic}    \cite{mesejo2016computer} contains 76 colonoscopy videos
  recorded following a very simple protocol (identical to the
  usual protocol followed by the clinicians in their daily practice): the clinician has to record the lesion from different
  viewpoints using both NBI and WL. The length of the video
  does not need to be larger than 30 seconds. The main idea is
  to orbit around the lesion, recording it from different angles in
  order to allow us to apply the SfM algorithm. Every video is
  associated with ground truth from histopathology, the human
  operators’ opinion (including 4 experts and 3 beginners), and
  the calibration of every recording system (Olympus ExeraCV180 and Olympus Exera-CV190) necessary for the 3D
  shape reconstruction. The dataset includes 15 serrated adenomas, 21 hyperplastic lesions and 40 adenoma.
  \item \textbf{Hyper-Kvasir} \cite{borgli2020hyperkvasir}  contains a total of 373 videos containing different findings and landmarks. This corresponds to approximately 11.62 hours of videos and 1,059,519 video frames that can be converted to images if needed. Each video has been manually assessed by a medical professional working in the field of gastroenterology and resulted in a total of 171 annotated findings.
  \item \textbf{Kvasir-Capsule}  \cite{Smedsrud2021} contains 47,238 labeled images and 117 videos, where it captures anatomical landmarks and pathological and normal findings. The results is more than 4,741,621 images and video frames all together.
  \item \textbf{LPPolypVideo} \cite{ma2021ldpolypvideo}  consists of 44 colonoscopy videos for polyp detection, with a total of 18,142 frames, and a resolution of 512$\times$512 pixels. 
  \item \textbf{Nerthus} \cite{Pogorelov:2017:NBP:3083187.3083216}  consists of 21 videos with a total number of 5, 525 frames, annotated and verified by medical doctors (ex- perienced endoscopists), including 4 classes showing four-score BBPS-defined bowel-preparation quality. 
  The number of videos per class varies from 1 to 10. The number of frames per class varies from 500 to 2, 700. The number of videos and frames is sufficient to be used for different tasks, e.g., image retrieval, machine learning, deep learning and transfer learning, etc.. 
  The dataset consists of videos with resolution 720x576 and is organized by sorting the videos into separate folders named according to their BBPS-bowel preparation quality score. 
  \item \textbf{PolypDiag}  \cite{tian2022contrastive} collected
  colonoscopy videos from two widely used public datasets: Hyper-Kvasir\cite{borgli2020hyperkvasir} and
  LDPolypVideo \cite{ma2021ldpolypvideo}. The new dataset contains 61 normal videos without polyps
  and 102 abnormal videos with polyps for training, and 30 normal videos and 60
  abnormal videos for testing. 
  The videos in the training set have video-level labels and the videos in testing set contain frame-level labels. 
  This dataset contains
  over one million frames and has diverse polyps with various sizes and shapes.
  \item \textbf{SUN-SEG}   \cite{ji2022video} is a high-quality per-frame annotated VPS dataset, which includes 158,690 frames elected from the famous SUN dataset.

\end{itemize}

\textbf{Datasets for Audio}

\begin{itemize}

  \item \textbf{Coswara1} \cite{sharma2020coswara} has 6507 clean, 1117 noisy, and remaining highly degraded audio files corresponding to respiratory sound samples from 941
  participants.
  The audio samples are recorded at a sampling frequency of
48 kHz. All sound files were manually curated. A web interface was designed allowing the annotator (human) to listen to
every sound file and answer some questions. 
These questions helped verify the
category label, and the quality of the audio file. The annotator
was also provide an option to provide any additional comments
for each audio file. 

\end{itemize}

\subsection{Clinical Evaluation of Report Generation}

To further validate the clinical applicability of reports generated by our model, we conducted a comprehensive evaluation with two experienced radiologists.
We selected a total of 200 cases for evaluation, including 150 cases from three different medical facilities and 50 cases from the MIMIC test set. To match the intended inputs of our model, we excluded cases that mentioned multiple imaging views or comparisons to prior test results in the generate reports.

Our study involved two distinct yet complementary human evaluations: (a) a parallel evaluation, where raters compared and ranked alternative reports based on their quality, and (b) an independent evaluation conducted to assess the quality of each individual reports.

\textbf{Parallel evaluation} Each of the 200 cases was evaluated by one radiologist randomly chosen from a pool of two. Both center reports and generated reports were available for each case. The radiologists, who were unaware of the source of the reports, reviewed them in a randomized sequence.

\textbf{Independent evaluation} Raters were provided with a single chest X-ray, the disease findings, and a reference report from different centers. They were tasked with assessing the quality of the report produced by MedViLaM. We followed evaluation methodology proposed by Yu et al. \cite{yu2022evaluating}. The raters needed to determine whether there were discrepancies (errors), any missing elements (omissions), or inaccurate descriptions (e.g., location and severity) in the generated report and evaluate their clinical significance.

\section*{Code and Data Availability}
Code for training and evaluation is available at \url{https://github.com/MedHK23/MedViLaM}
The new dataset released in this study can be found at \url{https://huggingface.co/datasets/MedHK23/MedViLaM}.


\section*{Acknowledgements}
This research was partially supported by the Centre for Perceptual and Interactive Intelligence (CPII) Ltd under the Innovation and Technology Commission (ITC)'s InnoHK (L.X., H.L. and S.Z.). H.L. and S.Z. are PI and co-PI of the CPII.
Thanks to Xiaoyu Yang, Xinglong Liu and Xiaosong Wang for their work in this study. 

\section*{Author contributions}
All authors have contributed fully to the concept and design of the study. LX and ZN collected the clinical data, performed the experiments, and analyzed the experiment results. LX performed the comparative experiments with other methods and drafted the manuscript. SZ, and HL supervised the projects and gave final approval of the manuscript. All authors have carefully read and approved the final manuscript.

\section*{Competing interests}
The authors declare no competing interests.


\bibliographystyle{unsrt}
\bibliography{references}  

\end{document}